\documentclass{article}

\usepackage{microtype}
\usepackage{graphicx}
\usepackage{subfigure}
\usepackage{booktabs} %

\usepackage{hyperref}

\usepackage[accepted]{icml2018}

\usepackage{amsmath}
\usepackage{amssymb}
\usepackage{macros}

\begin{document}

\twocolumn[
\icmltitle{Multi-task Maximum Causal Entropy Inverse Reinforcement Learning}

\icmlsetsymbol{equal}{*}

\begin{icmlauthorlist}
\icmlauthor{Adam Gleave}{berk}
\icmlauthor{Oliver Habryka}{berk}
\end{icmlauthorlist}

\icmlaffiliation{berk}{University of California, Berkeley}

\icmlcorrespondingauthor{Adam Gleave}{gleave@berkeley.edu}

\icmlkeywords{Inverse Reinforcement Learning, Multitask, Meta-Learning, Goal Specification}

\vskip 0.3in
]

\printAffiliationsAndNotice{}  %

\begin{abstract}
Multi-task Inverse Reinforcement Learning (IRL) is the problem of inferring multiple reward functions from expert demonstrations.
Prior work, built on Bayesian IRL, is unable to scale to complex environments due to computational constraints.
This paper contributes a formulation of multi-task IRL in the more computationally efficient Maximum Causal Entropy (MCE) IRL framework.
Experiments show our approach can perform one-shot imitation learning in a gridworld environment that single-task IRL algorithms need hundreds of demonstrations to solve.
We outline preliminary work using meta-learning to extend our method to the function approximator setting of modern MCE IRL algorithms.
Evaluating on multi-task variants of common simulated robotics benchmarks, we discover serious limitations of these IRL algorithms, and conclude with suggestions for further work.
\end{abstract}

\section{Introduction}
\begin{figure*}[h]
	\begin{center}
		\centerline{\includegraphics[width=\textwidth]{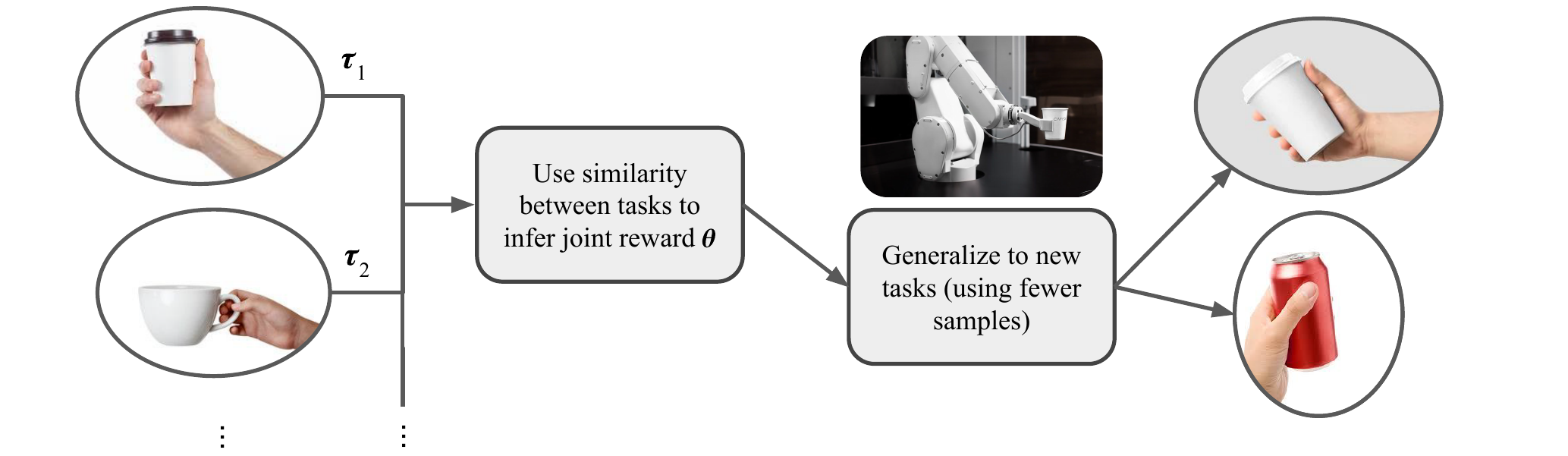}}
    \caption{Multi-task inverse reinforcement learning (IRL) uses demonstrations of similar tasks, such as grasping different types of containers, to jointly infer the reward functions for each task. 
    By exploiting the similarity between the reward functions, multi-task methods can achieve greater sample efficiency than conventional single-task IRL algorithms. 
    Multi-task methods can also rapidly generalise to new tasks, such as grasping a previously unseen type of container, with relatively few demonstrations.}
		\label{fig:intro} 
	\end{center}
	\vskip -0.2in
\end{figure*}
Inverse reinforcement learning (IRL) is the task of determining the reward function that generated a set of trajectories: sequences of state-action pairs~\cite{ng:2000}. 
It is a form of learning from demonstration that assumes the demonstrator follows a near-optimal policy with respect to an unknown reward. 
Sample efficiency is a key design goal, since it is costly to elicit human demonstrations.

In practice, demonstrations are often generated from multiple reward functions.
This situation naturally arises when the demonstrations are for different tasks, such as grasping different types of objects, as depicted in fig.~\ref{fig:intro}.
Less obviously, it also occurs when different individuals perform what is nominally the same task, reflecting individuals' unique preferences and styles. 
In this paper, we assume demonstrations of the same task are assigned a common label.

A naive solution to multi-task IRL is to repeatedly apply a single-task IRL algorithm to demonstrations of each task.
However, this method requires that the number of samples increases proportionally with the number of tasks, which is prohibitive in many settings.
Fortunately, the reward functions for related tasks are often similar, and exploiting this structure can enable greater sample efficiency.

Previous work on the multi-task IRL problem \cite{dimitrakakis:2011,babesvroman:2011,choi:2012} builds on Bayesian IRL \cite{ramachandran:2007}. 
Unfortunately, no extant Bayesian IRL methods scale to complex environments with high-dimensional, continuous state spaces such as robotics.
By contrast, approaches based on maximum causal entropy show more promise \cite{ziebart:2010}. 
Although the original maximum causal entropy IRL algorithm is limited to discrete state spaces, recent extensions such as guided cost learning and adversarial IRL scale to challenging continuous control environments \cite{finn:2016,fu:2018}.

Our two main contributions in this paper are:
\begin{itemize}
  \item \textbf{Regularised Maximum Causal Entropy (MCE)}. 
    We present a formulation of the multi-task IRL problem in the MCE framework.
	  Our approach simply adds a regularisation term to the loss, retaining the computational efficiency of the original MCE IRL algorithm.
    We evaluate in a $9 \times 9$ gridworld that takes hundreds of demonstrations for MCE IRL to solve.
    By contrast, after a single demonstration our regularised variant recovers a reward leading to a near-optimal policy.
	\item \textbf{Meta-Learning Rewards}.
	We describe preliminary work applying meta-learning to adversarial IRL.
  Evaluating on multi-task variants of continuous control tasks, we find baseline single-task adversarial IRL has poor performance even when given ample samples.
  This limitation consequently effects our multi-task variant.
	The poor performance appears to be due to the multimodal nature of optimal policies in these environments, presenting a challenge not present in other benchmarks.
	We conjecture this is analogous to the mode collapse problem in generative adversarial networks \cite{goodfellow:2014}, and conclude with suggestions for further work.
\end{itemize}

\section{Preliminaries and Single-Task IRL}
A Markov Decision Process (MDP) $M$ is a tuple $\mdp$ where 
$\statespace$ and $\actionspace$ are sets of states and actions;
$\transitiondist(s,a)(s')$ is the probability of transitioning to $\state'$ from $\state$ after taking action $\action$; $\discount \in [0,1]$ is a discount factor;
$\initialstatedist(\state)$ is the probability of starting in $\state$;
and $\reward(\state, \action)$ is the reward upon taking action $\action$ in state $\state$.
We write {\mdpnorliteral} to denote an MDP without a reward function.%

In the single-task IRL problem, the IRL algorithm is given access to an {\mdpnorliteral} and demonstrations $\demonstrations$ from an (approximately) optimal policy.
The goal is to recover a reward function $\reward$ that explains the demonstrations $\demonstrations$. 
Note this is an ill-posed problem: many reward functions $\reward$, including the constant zero reward function $\reward(\state,\action) = 0$, make the demonstrations $\demonstrations$ optimal.

Bayesian IRL addresses this identification problem by inferring a posterior distribution \cite{ramachandran:2007}. Although some probability mass will be placed on degenerate reward functions, for reasonable priors the majority of the probability will lie on more plausible explanations.

By contrast, maximum causal entropy chooses a single reward function, using the principle of maximum entropy to select the least specific reward function that is still consistent with the demonstrations \cite{ziebart:2008,ziebart:2010}. It models the demonstrations as being sampled from:
\begin{equation}
\label{eq:mce:policy}
\policy(\action \mid \state) = \exp\left(\qsoft(\state, \action) - \vsoft(\state, \action)\right),
\end{equation}
a stochastic expert policy that is noisily optimal for:
\begin{align}
\label{eq:mce:bellman}
\qsoft(\state, \action) &= \reward(\state, \action) + \discount \expectation_{\state' \sim T(s,a)}\left[V^{\text{soft}}(\state')\right], \\
\vsoft(\state) &= \softmax_{\action} \qsoft(\state, \action). \nonumber
\end{align}
Note there can exist multiple solutions to these softmax Bellman equations \cite{asadi:2017}. 

To reduce the dimension of the problem, it is common to assume the reward function is linear in features over the state-action pairs:
\begin{equation}
R(\state, \action) = \weights^T \featuremap(\state, \action).
\end{equation}
Let the expert demonstration $\demonstrations$ consist of $N$ trajectories $\demonstrations^{(j)} = \left(\state^{(j)}_0, \action^{(j)}_0, \dots, \state^{(j)}_T, \action^{(j)}_T\right)$. For convenience, write:
\begin{align}
\featuremap\left(\demonstrations^{(j)}\right) &= \sum_{t=0}^T \gamma^t \featuremap\left(\state^{(j)}_t, \action^{(j)}_t\right), \\
\featuremap(\tau) &= \frac{1}{N} \sum_{j=1}^N \featuremap\left(\demonstrations^{(j)}\right).
\end{align}
Given a known feature map $\featuremap$, the IRL problem reduces to finding weights $\weights$.

A key insight behind maximum causal entropy IRL is that actions in the trajectory sequence depend causally on previous states and actions: i.e.\ $\action^{(j)}_n$ may depend on $\state^{(j)}_0,\dots,\state^{(j)}_n$ and $\action^{(j)}_0,\dots,\action^{(j)}_{n-1}$, but not on states or actions that occur later in time. The causal log-likelihood of a trajectory $\tau^{(j)}$ is defined to be:
\begin{equation}
\mathcal{L}(\tau^{(j)}) = \sum_{t=0}^T \log \pr\left(\action^{(j)}_t \mid \state^{(j)}_{0:t}, \action^{(j)}_{0:t-1}\right),
\end{equation}
with the causal entropy of a policy defined in terms of the causal log-likelihood of its trajectories:
\begin{equation}
H(\pi) = \expectation_{\tau \sim \pi}\left[-L(\tau)\right].
\end{equation}
Maximum causal likelihood estimation of $\weights$ given the expert demonstrations $\demonstrations$ is equivalent to maximising the causal entropy of the stochastic policy $\policy$ subject to the constraint that its expected feature counts match those of the demonstrations:
\begin{equation}
F(\pi) \triangleq \expectation_{\substack{\state_0 \sim \initialstatedist \\ \action_{t+1} \sim \policy(\action_t \mid \state_t) \\ \state_{t+1} \sim \transitiondist(\state_t,\action_t)}} \sum_{t=0}^{\infty} \discount^t \featuremap(\state_t, \action_t) = \featuremap(\demonstrations).
\end{equation}
Note this constraint guarantees $\policy$ attains the same (expected) reward as the expert demonstrations \cite{abbeel:2004}.
Maximum causal entropy thus recovers reward weights that match the performance of the expert, while avoiding degeneracy by maximising the diversity of the policy.

\section{Methods for Multi-Task IRL}
In multi-task IRL, the reward $\reward_i$ varies between MDPs $M_i = \left(\statespace, \actionspace, \transitiondist, \discount, \initialstatedist, \reward_i\right)$ with associated expert demonstrations $\demonstrations_i$.
If the reward functions $\reward_i$ are unrelated to each other, we cannot do better than repeated application of a single-task IRL algorithm.
However, in practice similar tasks have reward functions with similar structure, enabling specialised multi-task IRL algorithms to accurately infer the reward with fewer demonstrations.

In the next section, we solve the multi-task IRL problem using the original maximum causal entropy IRL algorithm with an additional regularisation term. 
Following this, we describe how our method can be extended to scalable approximations of maximum causal entropy IRL.

\subsection{Regularised Maximum Causal Entropy IRL}

\label{sec:reg-mce:multi}
In the multi-task setting, we must jointly infer reward weights $\theta_i$ that explain each demonstration $\demonstrations_i$.
To make progress we must make some assumption on the relationship between different reward weights.
A natural assumption is that the reward weights for most tasks lie close to the mean across all tasks, i.e.\ $\lambda \|\theta_i - \bar{\theta}\|_2^2$ should be small, where $\lambda > 0$.
This corresponds to a prior that $\theta_i$ is drawn from i.i.d.\ Gaussians with mean $\bar{\theta}$ and variance monotonic with $\lambda$.
In practice, we do not know $\bar{\theta}$, but we can estimate it by taking the mean of the current iterates for $\theta_i$. 
This results in a pleasingly simple inference procedure. 
The regularised loss is:
\begin{equation}
\loss_i(\weights_i) = \sum_{j=1}^N \log P\left(\demonstrations^{(j)}_i\right)  + \frac{1}{2}\lambda\|\weights_i - \bar{\weights}\|_2^2,
\end{equation}
with gradient:
\begin{equation}
\nabla \loss_i(\theta_i)  = \featuremap(\demonstrations_i) - F(\pi) - \lambda\left(\weights_i - \bar{\weights}\right).
\end{equation}

\subsection{Meta-Learning Reward Networks}

\begin{algorithm}[b]
  \begin{algorithmic}
    \STATE Randomly initialise reward network parameters $\phi_0$
    \FOR{$t=1$ to $T$}
      \STATE Sample task $i$ with demonstrations $\tau_i$
      \STATE Set $\theta_0 \leftarrow \phi_{t - 1}$
      \FOR{$n=1$ to $N$}
        \STATE $\theta_n \leftarrow \mathrm{AIRL}(\theta_{n-1}, \tau_i)$, one step of adversarial IRL
      \ENDFOR
      \STATE Update $\phi_t \leftarrow \phi_{t-1} + \alpha(\theta_N - \phi_{t-1})$
    \ENDFOR
  \end{algorithmic}
  \caption{Meta-AIRL: Reptile and adversarial IRL}
  \label{algo:reptile-irl}
\end{algorithm}

In the previous section, we saw how multi-task IRL can be incorporated directly into the Maximum Causal Entropy (MCE) framework. 
However, the original MCE IRL algorithm has two major limitations.
First, it assumes the MDP's dynamics $\transitiondist$ are known, whereas in many applications (e.g.\ robotics) the dynamics are unknown and must also be learned.
Second, it requires the practitioner to provide a feature mapping $\featuremap$ such that the resulting reward $\reward$ is linear. 
For many tasks, finding these features may be the bulk of the problem, negating the benefit of IRL. 

Both of these shortcomings are addressed by guided cost learning \cite{finn:2016} and its successor adversarial IRL \cite{fu:2018}, scalable approximations of MCE IRL. 
Specifically, adversarial IRL uses a neural network to represent the reward $\reward_i$ as a function from states and actions, obviating the need to specify a feature map $\featuremap$.
Furthermore, it can handle unknown transition dynamics since it estimates the loss gradient via sampling rather than direct computation, and so only requires access to a simulation of the environment for rollouts.

Naively, we could directly translate the regularisation approach given in the previous section to this setting, applying it to the parameters $\theta_i$ of the neural network $R_i$.
However, regularising the parameter space may not regularise the output space: small changes in some parameters may have a large effect on the predicted reward, while large changes in other parameters may have little effect.

A more promising approach is to meta-learn the reward network parameters $\theta_i$.
We selected Reptile \cite{nichol:2018} as the basis for our initial experiments due to its computational efficiency, a key consideration given that IRL in complex environments is already computationally demanding.
Moreover, Reptile attains similar accuracy in few-shot supervised learning tasks as more computationally expensive algorithms.

Our meta adversarial IRL (meta-AIRL) method is described in algorithm \ref{algo:reptile-irl}.
We seek to find an initialisation $\phi$ for the reward network that can be quickly finetuned for any given task (by running adversarial IRL on demonstrations of that task).
To achieve this, we repeatedly sample a task and run $N$ steps of adversarial IRL, starting from our current initialisation $\phi$.
The initialisation is then updated along the line between the initialisation and final iterate of adversarial IRL.
Although this appears superficially similar to joint training, for $N \geq 2$ it is an approximation to first-order model-agnostic meta-learning (MAML) \cite{finn:2017}, a more principled but computationally expensive meta-learning algorithm.

Algorithm \ref{algo:reptile-irl} cannot be applied verbatim since adversarial IRL jointly learns a reward function \textit{and} a policy optimising that reward function. 
This is analogous to a GAN, where the policy network is a generator and the reward network defines a discriminator (assigning greater probability to higher reward trajectories).
We developed two concrete implementations of meta-AIRL, differing only in the policy used during meta-training:

\begin{itemize}
  \item \textbf{Random}.
  The simplest solution is to randomly initialise the policy at the start of each new task.
  This reduces adversarial IRL to a pure sample-based approximation of MCE IRL.
  We expect this to work in simple environments, where a random agent can cover most of the state space, but to fail in more challenging environments.

  \item \textbf{Task-specific}.
  A more sophisticated option is to maintain separate policy parameters per task.
  This method learns reward parameters that can be quickly finetuned to discriminate data from a distribution of generators.

  However, since the policy for a task is updated only when that task is sampled, care must be taken to ensure the frequency between samples does not grow too large. 
  Otherwise, policies for many tasks might become very suboptimal for the current reward network weights, slowing convergence.
  Accordingly, we suggest training in mini-batches of small numbers of tasks.
\end{itemize}

\section{Related Work}
Previous work in multi-task IRL has approached the problem from a Bayesian perspective. 
\citet{dimitrakakis:2011} model reward-policy function pairs $(\reward_i, \pi_i)$ as being drawn from a common (unknown) prior, over which they place a hyperprior.
This work provides a solid theoretical basis for work on multi-task IRL, but the inference problem is intractable even for moderately sized finite-state MDPs.

Complementary work has tackled an unlabelled variant of the multi-task IRL problem.
That is, not only are the reward functions $\reward_i$ unknown, it is also not known which reward $R_i$ each trajectory is paired with.
\citet{babesvroman:2011} use expectation-maximisation to cluster trajectories, an approach applicable to several IRL algorithms.
\citet{choi:2012} instead take a Bayesian IRL approach using a Dirichlet process mixture model, allowing a variable number of clusters.
Both methods reduce the problem to multiple single-task IRL problems, and so unlike our work do not exploit similarities between reward functions.

\citet{amin:2017} have studied the similar problem of repeated IRL: learning a common reward component shared across tasks, given known task-specific reward components. Although this could be solved by applying IRL to any one of the tasks, a repeated IRL algorithm can attain better bounds and even resolve the ambiguity inherent in single-task IRL.

IRL is often used for imitation learning.
Multi-task imitation learning has also been studied from a non-IRL perspective, especially in the context of generative adversarial imitation learning (GAIL) \cite{ho:2016}.
Recent extensions to GAIL augment trajectories with a latent intention variable that specifies the task, and then maximise mutual information between the state-action pairs and intention variable \cite{hausman:2017,li:2017}.
These approaches are focused on disentangling trajectories from different tasks, and are not intended to speed up the learning of \textit{new} tasks.

However, the imitation learning community does address this problem in one-shot imitation learning: having seen a distribution of trajectories over various tasks, learn a new task from a single demonstration.
\citet{ziyuwang:2017} use GAIL with the discriminator conditioned on the output generated by an LSTM encoder.
After training on unlabelled trajectories, this method can perform one-shot imitation learning by conditioning on the code of a demonstration trajectory.
One-shot imitation learning has also been tackled within the behavioural cloning paradigm \cite{duan:2017}.

Multi-agent GAIL \cite{song:2018} is the imitation learning method most similar to our paper.
Although GAIL does not explicitly learn a reward function, it is equivalent to IRL composed with RL.
Similar to our work, multi-agent GAIL seeks to improve sample efficiency by exploiting similarity between the reward functions.
However, unlike our work, multi-agent GAIL makes strong assumptions on the reward function (e.g.\ zero sum games).

In concurrent work, \citet{xu:2018} independently developed a meta-IRL algorithm, applying MAML directly to Maximum Entropy IRL.
They obtain good performance on a ``Spriteworld'' navigation domain, consisting of a gridworld with overlaid ``sprite'' textures.
However, their algorithm inherits the limitations of Maximum Entropy IRL: the MDP must have a finite state space and known transition dynamics.
By contrast, our meta-AIRL method (algorithm~\ref{algo:reptile-irl}) learns the transition dynamics and can operate in infinite state space MDPs such as continuous control environments.
However, this flexibility comes at a cost, with our experiments showing that adversarial IRL has difficulty learning from non-unimodal policies.
Accordingly, we view \citet{xu:2018} and our own work as being complementary, making different trade-offs in order to target different applications.

\begin{figure}
	\vskip -0.1in
	\begin{center}
		\centerline{\includegraphics{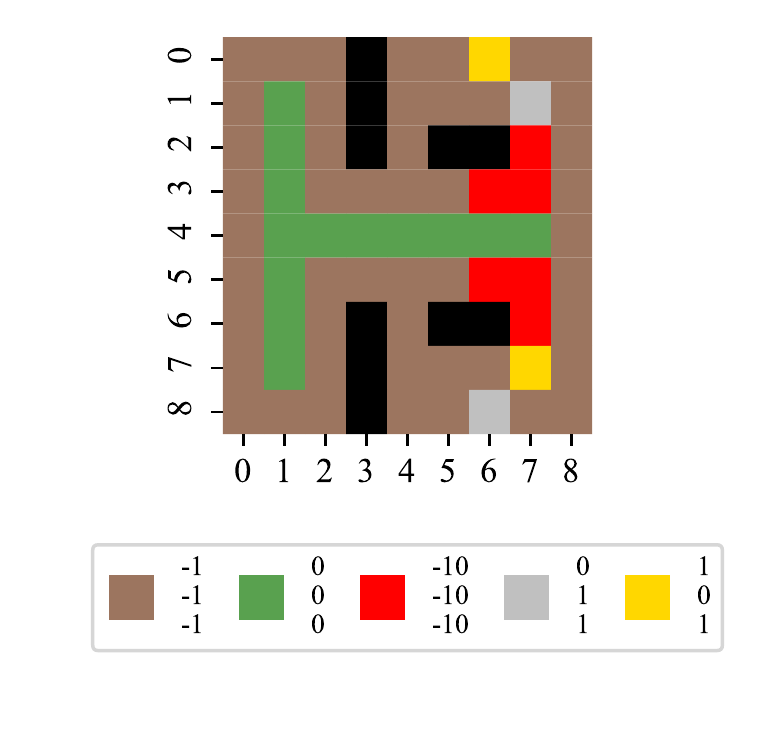}}
		\vskip -0.2in
		\caption{A depiction of the $9\times 9$ gridworld we evaluate on. Black cells denote walls that are impassable. All other colour cells denote a particular feature, assigned the reward specified in the legend. 
			The top row gives the weight for reward function \texttt{A}, the middle row for \texttt{B} and the final row for \texttt{A+B}. 
			Transitions are stochastic, with probability $0.8$ of moving in the desired action, and $0.1$ of moving in each of the two orthogonal directions.}
		\label{fig:experiments:jungle:gt} 
	\end{center}
	\vskip -0.2in
\end{figure}

\section{Experiments}

\subsection{regularised Maximum Causal Entropy IRL}

We evaluate our regularised maximum causal entropy (MCE) IRL algorithm in a few-shot reward learning problem on the gridworld depicted in fig.~\ref{fig:experiments:jungle:gt}.
Transitions in the gridworld are stochastic, with probability $0.8$ of moving in the desired direction, and $0.1$ of moving in each of the two orthogonal directions.
Each cell in the gridworld is either a wall (in which case the state can never be visited), or one of five objects types: dirt, grass, lava, gold and silver.

We define three different reward functions \texttt{A}, \texttt{B} and \texttt{A+B} in terms of these object types, as specified by the legend of fig.~\ref{fig:experiments:jungle:gt}.
The reward functions assign the same weights to dirt, grass and lava, but differ in the weights for gold and silver.
\texttt{A} likes silver but is neutral about gold, \texttt{B} has the opposite preferences and \texttt{A+B} likes both gold and silver.
We generate synthetic demonstrations for each of these three reward functions using the MCE planner given by eq.~\eqref{eq:mce:policy}.

Our multi-task IRL algorithm is then presented with demonstrations from an optimal policy for each reward function.
Demonstrations for the few-shot environment are restricted to $M$ trajectories, varying between $1$ and $100$, while demonstrations for the other two environments contain $N = 1000$ trajectories.
To make the task more challenging, our algorithm is not provided with the feature representation, instead having to learn the reward separately for each state. 
We repeat all experiments for $5$ random seeds.

\subsubsection{Comparison to Baselines}
\begin{figure*}[p]
	\begin{center}
		\centerline{\includegraphics[width=\textwidth]{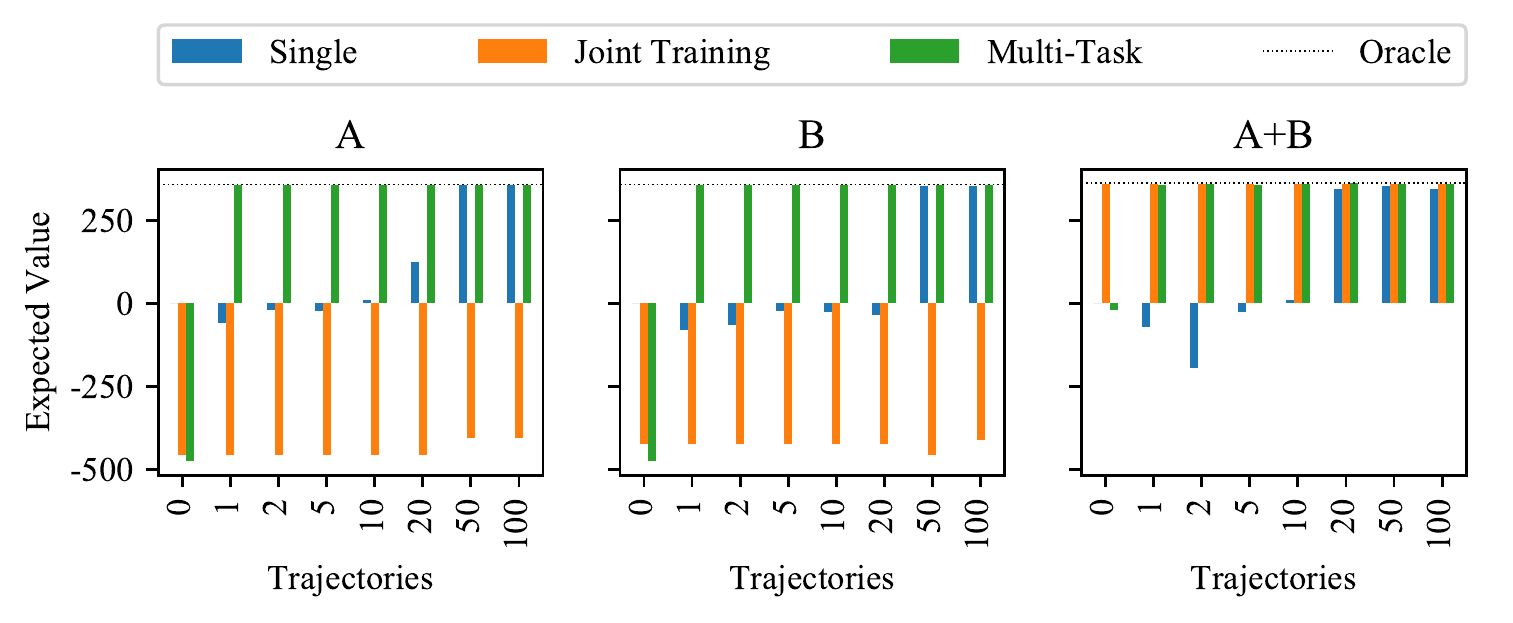}}
		\caption{Value obtained from an optimal policy for the inferred reward, by the number of trajectories observed ($x$-axis) and algorithm (colour). Values reported are the best out of 5 random seeds. The oracle (dashed horizontal line) is the value obtained by an optimal policy on the ground truth reward.}
		\label{fig:experiments:jungle:baseline-comparison}
	\end{center}
\end{figure*}

We compare against two baselines.
The first (`single') corresponds to using single-task MCE IRL, seeing only the $M$ trajectories from the few-shot environment.
The second (`joint training') combines the demonstrations from all three environments into a single $2N+M$-length sequence of trajectories.
For reference, we also display the value obtained by an optimal (`oracle') policy. %
Figure~\ref{fig:experiments:jungle:baseline-comparison} shows the best out of 5 random seeds.%

Our multi-task IRL algorithm recovers a near-optimal policy in all 5 runs after only two trajectories, and in the best case requires only a single trajectory.
By contrast, the `single' baseline requires $M=50$ trajectories or more to recover a good policy even in the best case, and after $100$ trajectories several seeds still obtain negative total rewards.

The `joint training' baseline performs well on \texttt{A+B}.
This is unsurprising, since an optimal policy in \texttt{A} or \texttt{B} is near-optimal in \texttt{A+B}.
However, it fares poorly in both the \texttt{A} and \texttt{B} environments, never obtaining a positive reward even in the best case.

Note that all approaches fail in the zero-shot case on \texttt{A} and \texttt{B}, making the success of multi-task IRL in the few-shot case all the more remarkable.
Demonstrations solely from non-target environments are not enough to recover a good reward in the target, and so substantial learning must be taking place with only one or two trajectories.

\subsubsection{Hyperparameter Choice}

\begin{figure*}[p]
	\vskip 0.2in
	\begin{center}
		\centerline{\includegraphics[width=\textwidth]{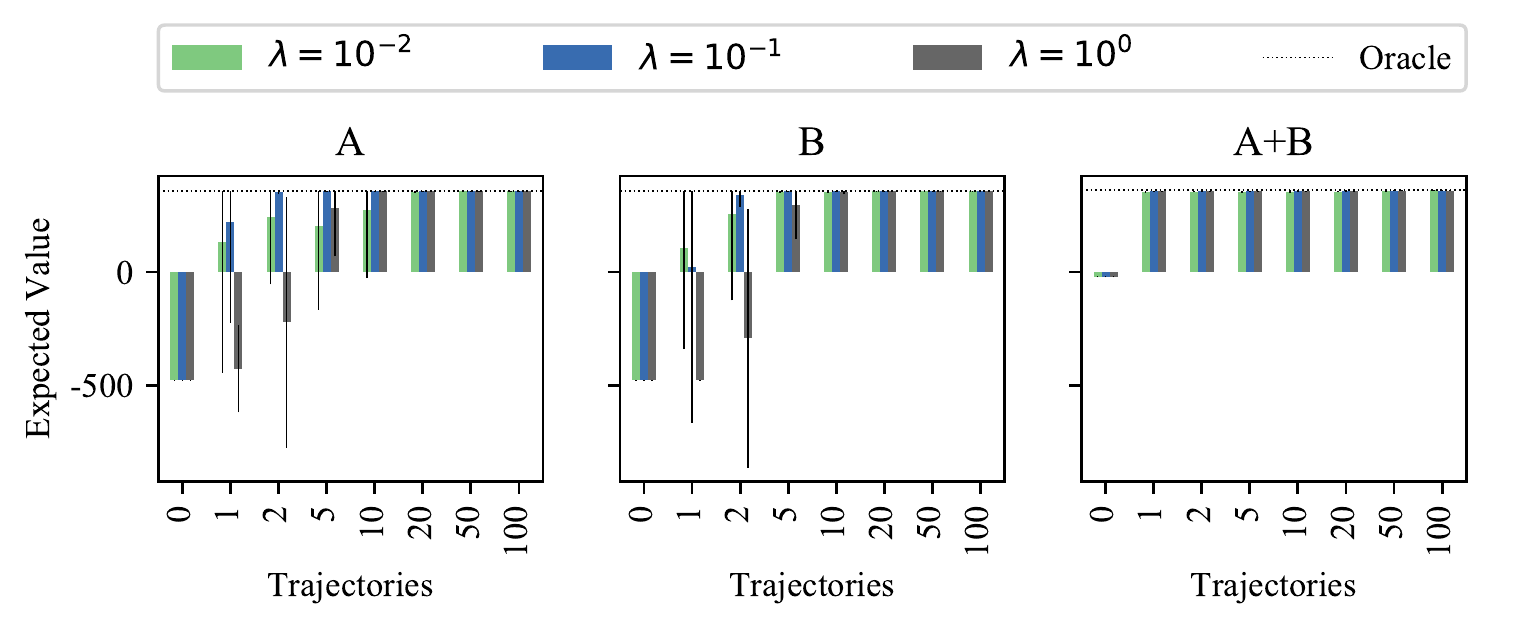}}
		\caption{Expected value obtained from an optimal policy for the inferred reward, by the number of trajectories observed ($x$-axis) and regularisation constant $\lambda$ (colour).
			Solid bars represent the mean and error bars span the 95\% confidence interval computed from 5 random seeds.
			The oracle (dashed horizontal line) is the value obtained by an optimal policy on the ground truth reward.}
		\label{fig:experiments:jungle:reg-comparison}
	\end{center}
	\vskip -0.2in
\end{figure*}

\label{appendix:regularised-mce:hyperparameters}
Our regularised MCE IRL algorithm takes a hyperparameter $\lambda$ that specifies the regularisation strength.
We show in fig.~\ref{fig:experiments:jungle:reg-comparison} the results of a hyperparameter sweep between $\lambda = 10^{-2}$ and $\lambda = 10^0$.
As expected, the weakest regularisation constant $\lambda = 10^{-2}$ suffers from high variance across the random seeds when the number of trajectories is small.

Perhaps more surprising, the strongest regularisation constant $\lambda = 10^0$ also has high variance.
We conjecture that it imposes too strong a prior, making it highly sensitive to the trajectories observed in the off-target environments.

The median regularisation constant $\lambda = 10^{-1}$ attains the lowest variance and highest mean of the hyperparameters tested, and was used in the previous section's experiments.

\begin{figure*}
	\begin{center}
		\vskip -0.1in
		\centerline{\includegraphics{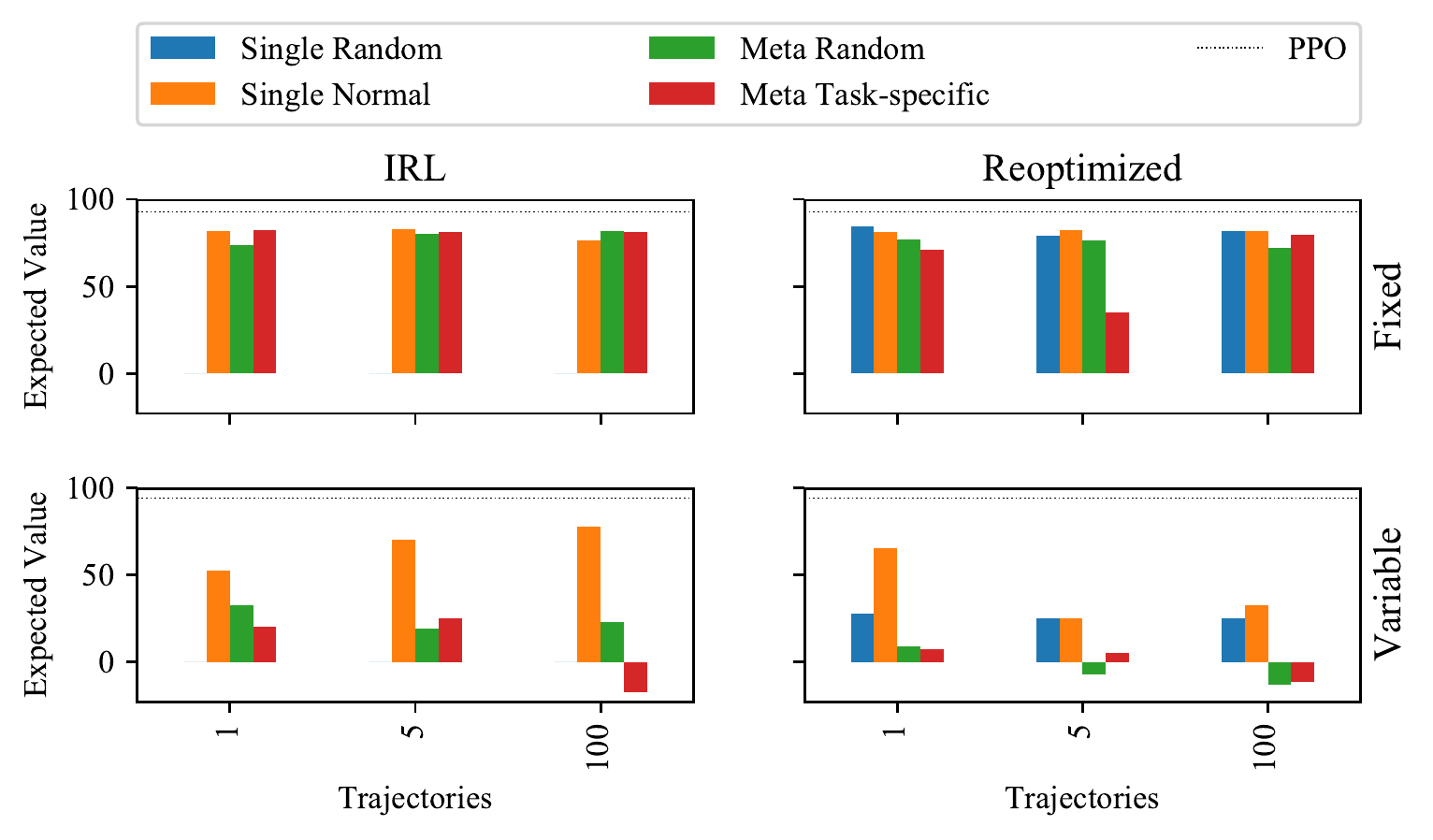}}
		\caption{Policy value. Left: IRL policy jointly learnt with reward; right: policy from training PPO using the IRL reward. Top: fixed test case, meta-training on left and right and finetune on left; bottom: variable test case, meta-training on red and blue and finetune on blue. Values are grouped by the number of finetuning trajectories ($x$-axis) and algorithm (colour), and are the best out of 5 random seeds. The oracle (dashed horizontal line) is the value obtained by the expert PPO policy.}
		\label{fig:experiments:mountaincar}
		\vskip -0.3in
	\end{center}
\end{figure*}

\begin{figure}
	\begin{center}
		\centerline{\includegraphics[width=0.5\columnwidth]{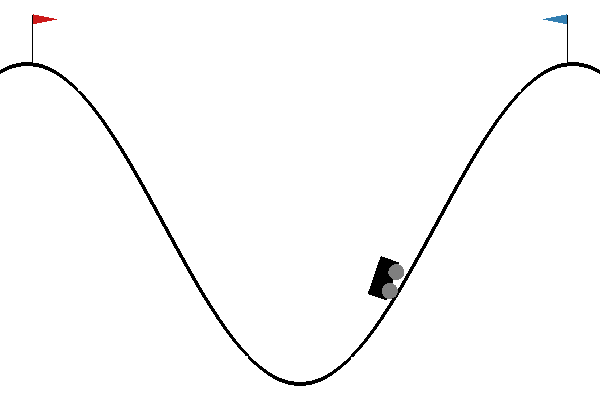}}
		\caption{Still image from our symmetric variant of Gym's \texttt{MountainCarContinuous-v0} environment.}
		\label{fig:experiments:mountaincar-env}
	\end{center}
	\vskip -0.3in
\end{figure}
These results indicate that where sample efficiency is paramount, it is important to choose a regularisation hyperparameter suited to the task distribution.
However, the algorithm is reasonably robust to hyperparameter choice, with all parameters (varying across two orders of magnitude) attaining near-optimal performance after as few as 20 trajectories.
By contrast, the single-task IRL algorithm did not achieve this level of performance even in the best case until observing 50 or more trajectories.

\subsection{Meta-Learning Reward Networks v2}

We evaluated both our \textit{random} and \textit{task-specific} implementations of meta-AIRL (algorithm~\ref{algo:reptile-irl}) on a multi-task variant of the mountain car continuous control problem.
As a baseline we compare against both a \textit{standard} version of single-task AIRL and one with a \textit{random} policy.

Our test environment, illustrated in figure~\ref{fig:experiments:mountaincar-env}, is a symmetric version of Gym's \texttt{MountainCarContinuous-v0} environment.
Both the left and right mountains have a flag at their peak, and the episode ends as soon as the car touches either flag.
One flag is the \textit{goal}, and the agent receives $100$ reward from reaching this flag.
The other flag is a \textit{decoy}, and the agent receives a $100$ penalty if it touches this flag.
In addition, there is a quadratic control cost.

We evaluate in two test cases. The \textit{fixed} test case consists of two environments: one where the goal flag is always on the left, another where it is always on the right.
In the \textit{variable} test case, the side of the goal flag is chosen randomly at the start of each episode.
It consists of two environments: one where the blue flag is the goal, another where it is the red flag.
The position of both flags is included in the state.

Figure~\ref{fig:experiments:mountaincar} reports the value of the resulting policies in the fixed test case (top row) and variable test case (bottom row).
The left column corresponds to policies learnt directly by AIRL, and the right column policies trained with PPO on the reward learnt by AIRL.

For the fixed test case, we see that both the single-task baseline and our meta-AIRL algorithm produce near-optimal solutions.
This is unsurprising: the optimal policy is unimodal, and so it is simple to extrapolate from a single trajectory, especially in a low-dimensional environment such as mountain car.

In the variable test case, figure~\ref{fig:experiments:mountaincar} shows that single-task AIRL is unable to reliably find a good solution even after observing $100$ trajectories.
Reptile can only learn a good meta-initialisation in the outer loop if consistent progress is made in the AIRL inner loop, so unsurprisingly our meta-AIRL algorithm also fails in this environment.
Note the variable test case has a bimodal expert policy: the best trajectory depends on whether the target flag is on the left or right.

Our findings suggest that adversarial IRL attains good performance only in environments with a unimodal optimal policy. In these environments, a handful of trajectories is sufficient to recover the reward, leaving little room for improvement from applying meta-learning. While existing simulated robotics benchmarks can largely be solved by unimodal policies, many practical tasks (such as multi-step assembly) cannot, making this a pressing area for further research.

\section{Conclusions and Future Work}
Sample efficient solutions to the multi-task IRL problem are critical for enabling real-world applications, where collecting human demonstrations is expensive and slow.
The multi-task IRL problem has previously been studied exclusively from a Bayesian IRL perspective.
In this paper we took the alternative approach of formulating the multi-task problem inside the maximum causal entropy IRL framework by adding a regularisation term to the loss.
Experiments find our multi-task IRL algorithm can perform one-shot imitation learning in an environment that single-task IRL requires hundreds of demonstrations to learn.

Maximum causal entropy IRL \cite{ziebart:2010} cannot scale to MDPs with large or infinite state spaces, and moreover requires known dynamics.
Both these problems have been alleviated by recent extensions to maximum causal entropy IRL, such as guided cost learning and adversarial IRL \cite{finn:2016,fu:2018}. 
Our second contribution is to show how in this function approximator setting, multi-task IRL can be framed as a meta-learning problem.

Testing of our prototype meta-AIRL method (algorithm~\ref{algo:reptile-irl}) found that adversarial IRL~\cite{fu:2018} can only learn from unimodal expert policies, seriously limiting the applicability of meta-AIRL.
We conjecture this limitation in adversarial IRL is related to the well-known problem of mode collapse in generative adversarial networks (GAN). 
A fruitful research direction might be to apply recent innovations in GAN training such as unrolling the optimisation of the discriminator \cite{metz:2017} or variational learning \cite{srivastava:2017} to stabilise adversarial IRL training. 

Another limitation of adversarial IRL is that the inferred reward network (discriminator) often overfits to the jointly learnt policy (generator).
In particular, training a policy using the reward network fails from most random initialisations, even if the policy learnt by adversarial IRL obtains good performance.
Our prototype performed meta-learning only on the reward network, making it particularly sensitive to this limitation.
Jointly meta-learning the reward and policy network could improve performance, but we believe this will first require significant improvements in meta-RL algorithms.

The source code for our algorithms and experiments is open source and available at 
\ifdefined\isaccepted
\url{https://github.com/HumanCompatibleAI/population-irl}
\else
---removed for double blind---
\fi
.

\section*{Acknowledgements}
\ifdefined\isaccepted
Thanks to Rohin Shah and Stuart Russell for productive discussions; Jennifer Chen for creating several of the figures; S\"{o}ren Mindermann, Kelvin Xu and Daniel Filan for feedback on earlier versions of this paper; and the anonymous reviewers for their comments.
\else
Removed for double blind review.
\fi

\bibliography{refs}
\bibliographystyle{icml2018}

\end{document}